\documentclass{article} 

\makeatletter  

\usepackage{fontspec}

\usepackage{iftex}   
\ifXeTeX
  \usepackage{fontspec}
  \usepackage{xeCJK}
  \setCJKmonofont[Path=02-Font/,Extension=.ttf]{LiberationMono}
\else
\fi

\usepackage[verbose=true,letterpaper]{geometry}
\AtBeginDocument{
  \newgeometry{
    textheight=9in,
    textwidth=6.5in,
    top=1in,
    headheight=14pt,
    headsep=25pt,
    footskip=30pt
  }
}

\newcommand{\headeright}{\textcolor{gray!60}{A Preprint}}
\newcommand{\undertitle}{A Preprint}
\newcommand{\shorttitle}{\@title}

\usepackage{fancyhdr}
\fancyheadoffset{0pt}
\fancypagestyle{plain}{%
  \fancyhf{} 
  
  \cfoot{\thepage}}

\def\keywordname{{\bfseries \emph{Keywords}}}%
\def\keywords#1{\par\addvspace\medskipamount{\rightskip=0pt plus1cm
\def\and{\ifhmode\unskip\nobreak\fi\ $\cdot$
}\noindent\keywordname\enspace\ignorespaces#1\par}}

\renewcommand{\normalsize}{%
  \@setfontsize\normalsize\@xpt\@xipt
  \abovedisplayskip      7\p@ \@plus 2\p@ \@minus 5\p@
  \abovedisplayshortskip \z@ \@plus 3\p@
  \belowdisplayskip      \abovedisplayskip
  \belowdisplayshortskip 4\p@ \@plus 3\p@ \@minus 3\p@
}
\normalsize
\renewcommand{\small}{%
  \@setfontsize\small\@ixpt\@xpt
  \abovedisplayskip      6\p@ \@plus 1.5\p@ \@minus 4\p@
  \abovedisplayshortskip \z@  \@plus 2\p@
  \belowdisplayskip      \abovedisplayskip
  \belowdisplayshortskip 3\p@ \@plus 2\p@   \@minus 2\p@
}
\renewcommand{\footnotesize}{\@setfontsize\footnotesize\@ixpt\@xpt}
\renewcommand{\scriptsize}{\@setfontsize\scriptsize\@viipt\@viiipt}
\renewcommand{\tiny}{\@setfontsize\tiny\@vipt\@viipt}
\renewcommand{\large}{\@setfontsize\large\@xiipt{14}}
\renewcommand{\Large}{\@setfontsize\Large\@xivpt{16}}
\renewcommand{\LARGE}{\@setfontsize\LARGE\@xviipt{20}}
\renewcommand{\huge}{\@setfontsize\huge\@xxpt{23}}
\renewcommand{\Huge}{\@setfontsize\Huge\@xxvpt{28}}

\providecommand{\section}{}
\renewcommand{\section}{%
  \@startsection{section}{1}{\z@}%
                {-2.0ex \@plus -0.5ex \@minus -0.2ex}%
                { 1.5ex \@plus  0.3ex \@minus  0.2ex}%
                {\large\bf\raggedright}%
}
\providecommand{\subsection}{}
\renewcommand{\subsection}{%
  \@startsection{subsection}{2}{\z@}%
                {-1.8ex \@plus -0.5ex \@minus -0.2ex}%
                { 0.8ex \@plus  0.2ex}%
                {\normalsize\bf\raggedright}%
}
\providecommand{\subsubsection}{}
\renewcommand{\subsubsection}{%
  \@startsection{subsubsection}{3}{\z@}%
                {-1.5ex \@plus -0.5ex \@minus -0.2ex}%
                { 0.5ex \@plus  0.2ex}%
                {\normalsize\bf\raggedright}%
}
\providecommand{\paragraph}{}
\renewcommand{\paragraph}{%
  \@startsection{paragraph}{4}{\z@}%
                {1.5ex \@plus 0.5ex \@minus 0.2ex}%
                {-1em}%
                {\normalsize\bf}%
}
\providecommand{\subparagraph}{}
\renewcommand{\subparagraph}{%
  \@startsection{subparagraph}{5}{\z@}%
                {1.5ex \@plus 0.5ex \@minus 0.2ex}%
                {-1em}%
                {\normalsize\bf}%
}

\newlength{\@abovecaptionskip}
\newlength{\@belowcaptionskip}

\renewenvironment{table}
  {\setlength{\abovecaptionskip}{\@belowcaptionskip}%
   \setlength{\belowcaptionskip}{\@abovecaptionskip}%
   \@float{table}}
  {\end@float}

\renewcommand{\footnoterule}{\kern-3\p@ \hrule width 12pc \kern 2.6\p@}
\def\@listi  {\leftmargin\leftmargini}
\def\@listii {\leftmargin\leftmarginii
              \labelwidth\leftmarginii
              \advance\labelwidth-\labelsep
              \topsep  2\p@ \@plus 1\p@    \@minus 0.5\p@
              \parsep  1\p@ \@plus 0.5\p@ \@minus 0.5\p@
              \itemsep \parsep}
\def\@listiii{\leftmargin\leftmarginiii
              \labelwidth\leftmarginiii
              \advance\labelwidth-\labelsep
              \topsep    1\p@ \@plus 0.5\p@ \@minus 0.5\p@
              \parsep    \z@
              \partopsep 0.5\p@ \@plus 0\p@ \@minus 0.5\p@
              \itemsep \topsep}
\def\@listiv {\leftmargin\leftmarginiv
              \labelwidth\leftmarginiv
              \advance\labelwidth-\labelsep}
\def\@listv  {\leftmargin\leftmarginv
              \labelwidth\leftmarginv
              \advance\labelwidth-\labelsep}
\def\@listvi {\leftmargin\leftmarginvi
              \labelwidth\leftmarginvi
              \advance\labelwidth-\labelsep}

\providecommand{\maketitle}{}
\renewcommand{\maketitle}{%
  \par
  \begingroup
    \renewcommand{\thefootnote}{\fnsymbol{footnote}}
    \long\def\@makefntext##1{%
      \parindent 1em\noindent
      \hbox to 1.8em{\hss $\m@th ^{\@thefnmark}$}##1
    }
    \thispagestyle{empty}
    \@maketitle
    \@thanks
  \endgroup
  \let\maketitle\relax
  \let\thanks\relax
}

\newcommand{\@toptitlebar}{
  \hrule height 2\p@
  \vskip 0.25in
  \vskip -\parskip%
}
\newcommand{\@bottomtitlebar}{
  \vskip 0.29in
  \vskip -\parskip
  \hrule height 2\p@
  \vskip 0.09in%
}

\providecommand{\@maketitle}{}
\renewcommand{\@maketitle}{%
  \vbox{%
    \hsize\textwidth
    \linewidth\hsize
    \vskip 0.1in
    \@toptitlebar
    \centering
    {\LARGE\bfseries \@title\par}
    \@bottomtitlebar
    {\small\MakeUppercase{\undertitle}}\\
    \vskip 0.1in
    \def\And{%
      \end{tabular}\hfil\linebreak[0]\hfil%
      \begin{tabular}[t]{c}\bf\rule{\z@}{24\p@}\ignorespaces%
    }
    \def\AND{%
      \end{tabular}\hfil\linebreak[4]\hfil%
      \begin{tabular}[t]{c}\bf\rule{\z@}{24\p@}\ignorespaces%
    }
    \begin{tabular}[t]{c}\bf\rule{\z@}{24\p@}\@author\end{tabular}%
  \vskip 0.4in \@minus 0.1in \center{\@date}   \vskip 0.2in
  }
}

\renewenvironment{abstract}
{
  \centerline
  {\large \bfseries Abstract}
  \begin{quote}
}
{
  \end{quote}
}

\makeatother  

\usepackage{amsmath}                
\usepackage{amssymb}                
\usepackage{amsfonts}               
\usepackage[protrusion=true,expansion=false,tracking=false]{microtype} 
\usepackage{latexsym}               

\usepackage{titlesec}               
\usepackage{ulem}                   
\usepackage{enumitem}               
\usepackage{verbatim}               
\usepackage{mdframed}               
\usepackage{pifont}                 
\newcommand{\xmark}{\ding{55}}      

\usepackage{array}                  
\usepackage{makecell}               
\renewcommand{\arraystretch}{1.5}   
\usepackage{booktabs}               

\usepackage{graphicx}               
\usepackage{float}                  
\usepackage{caption}                
\usepackage{subcaption}             

\usepackage{gensymb}                
\usepackage{siunitx}                
\DeclareSIUnit\angstrom{\text{Å}}   

\usepackage{hyperref}
\usepackage{xurl}                   

\hypersetup{
    colorlinks=true,
    linkcolor=blue,
    filecolor=magenta,
    urlcolor=blue,
    citecolor=blue,
    pdfauthor={Your Name},
    pdftitle={Your Paper Title},
}
\usepackage{cleveref}               

\usepackage[numbers,sort&compress]{natbib}
\usepackage{algorithm}              
\usepackage{algorithmic}            
\usepackage{xcolor}                 

\usepackage{amssymb}                 
\usepackage{xcolor}                  
\usepackage{tgheros}                 
\usepackage{sansmath}                
\usepackage{listings}                

\definecolor{color_ans_title}{RGB}{10,80,20}
\definecolor{color_ans_content}{RGB}{10,40,0}

\ifdefined\answerpart\else
\newenvironment{answerpart}[1]
{%
  \vspace{-0.0cm}
  \subsection*{{\fontfamily{qhv}\selectfont\textcolor{color_ans_title}{#1}}}%
  \vspace{-0.3cm}%
  \begingroup 
  \fontfamily{qhv}\selectfont\color{color_ans_content}%
}
{%
  \endgroup 
  \vspace{0.5cm}
}
\fi

\definecolor{citation}{rgb}{0.95,0.6,0.4}           
\definecolor{codegreen}{rgb}{0,0.6,0}               
\definecolor{codegray}{rgb}{0.5,0.5,0.5}            
\definecolor{codepurple}{rgb}{0.58,0,0.82}          
\definecolor{backcolour}{rgb}{0.95,0.95,0.92}       
\definecolor{codeblue}{rgb}{0,0,1}                  
\definecolor{codered}{rgb}{1,0,0}                   
\definecolor{magentacolor}{rgb}{1,0,1}              

\lstdefinelanguage{JSON}{
    basicstyle=\ttfamily\small,
    commentstyle=\color{codegreen},                 
    stringstyle=\color{codepurple},                 
    keywords={true,false,null},                     
    keywordstyle=\color{magenta},                   
    ndkeywords={},                                  
    ndkeywordstyle=\color{codeblue},                
    identifierstyle=\ttfamily,
    sensitive=false,
    breaklines=true,
    morestring=[b]",                                
    morestring=[s]{>}{<}                            
}

\lstdefinestyle{mystyle}{
    backgroundcolor=\color{backcolour},             
    commentstyle=\color{codegreen},                 
    keywordstyle=\color{magenta},                   
    numberstyle=\tiny\color{codegray},              
    stringstyle=\color{codepurple},                 
    basicstyle=\ttfamily\footnotesize,              
    breakatwhitespace=false,
    breaklines=true,
    captionpos=b,
    keepspaces=true,
    numbers=left,
    numbersep=5pt,
    showspaces=false,
    showstringspaces=false,
    showtabs=false,
    tabsize=2,
    keywords={GRI, Summary, point, source, tone, value, energy, emissions, scope,
              renewable, consumption, carbon, reduction, target, total, data,
              positive, negative, neutral, Scope1, Scope2, Scope3, GWh, metric},
    morekeywords={GRI_302_1, emissions_1, emissions_2, resource_1, resource_2,
                  biodiversity_1, environment_1, ems_1, ContentExtraction,
                  ToneAnalysis, NumericalData},
    emph={Summary,point,source,tone,value},
    emphstyle=\color{blue}
}
\title{ESGLens: An LLM-Based RAG Framework for Interactive ESG Report Analysis and Score Prediction}

\author{
    Tsung-Yu Yang \\
    Massachusetts Institute of Technology \\
    \texttt{tyyang@mit.edu}
    \And
    Meng-Chi Chen \\
    Massachusetts Institute of Technology \\
    \texttt{edchen93@mit.edu}
}

\date{\today}

\begin{document}
\maketitle

\begin{abstract}
Environmental, Social, and Governance (ESG) reports are central to investment decision-making, yet their length, heterogeneous content, and lack of standardized structure make manual analysis costly and inconsistent. We present ESGLens, a proof-of-concept framework combining retrieval-augmented generation (RAG) with prompt-engineered extraction to automate three tasks: (1)~structured information extraction guided by Global Reporting Initiative (GRI) standards, (2)~interactive question-answering with source traceability, and (3)~ESG score prediction via regression on LLM-generated embeddings. ESGLens is purpose-built for the domain: a report-processing module segments heterogeneous PDF content into typed chunks (text, tables, charts); a GRI-guided extraction module retrieves and synthesizes information aligned with specific standards; and a scoring module embeds extracted summaries and feeds them to a regression model trained against London Stock Exchange Group (LSEG) reference scores. We evaluate the framework on approximately 300 reports from companies in the QQQ, S\&P~500, and Russell~1000 indices (fiscal year 2022). Among three embedding methods (ChatGPT, BERT, RoBERTa) and two regressors (Neural Network, LightGBM), ChatGPT embeddings with a Neural Network achieve a Pearson correlation of 0.48 ($R^{2} \approx 0.23$) against LSEG ground-truth scores -- a modest but statistically meaningful signal given the ${\sim}300$-report training set and restriction to the environmental pillar. A traceability audit shows that 8 of 10 extracted claims verify against the source document, with two failures attributable to few-shot example leakage. We discuss limitations including dataset size and restriction to environmental indicators, and release the code to support reproducibility.\end{abstract}

\section{Introduction}

As climate change intensifies, various sectors are urging corporations to pursue purposes beyond shareholders' interests. One major initiative is the increasing publication of ESG reports, where many corporations disclose their efforts to mitigate climate change to investors, governments, academics, and advocates. \cite{byrne_esg_history, krantz_esg_history, wang2023environmental, bose2020evolution, rouen2023evolution} However, ESG reports are self-disclosed documents that lack a standardized template, making cross-referencing between companies and industries difficult \cite{cort2020esg, gobel_esg_unstandardized_2022, ecoactive_esg_challenges_2025}. In this research, we aim to leverage large language models to analyze ESG reports and enable users to ask targeted questions about specific reports.

In addition to analyzing ESG reports, another key motivation for this research is to extend the capability of LLMs to answer questions based on user-provided documents \cite{ding2025euleresgautomatingesgdisclosure, khan2024developingretrievalaugmentedgeneration}. Documents, such as ESG reports, are typically in PDF format and contain text, charts, and figures \cite{ding2025euleresgautomatingesgdisclosure}. Developing an efficient method and framework for document embedding and summarization would allow users to cross-reference multiple documents and ask more detailed questions \cite{yang2025curiousllmelevatingmultidocumentquestion}. This would be a valuable tool in various fields \cite{anaraki2025large}.

\begin{figure*}
\centering
\includegraphics[width=16 cm]{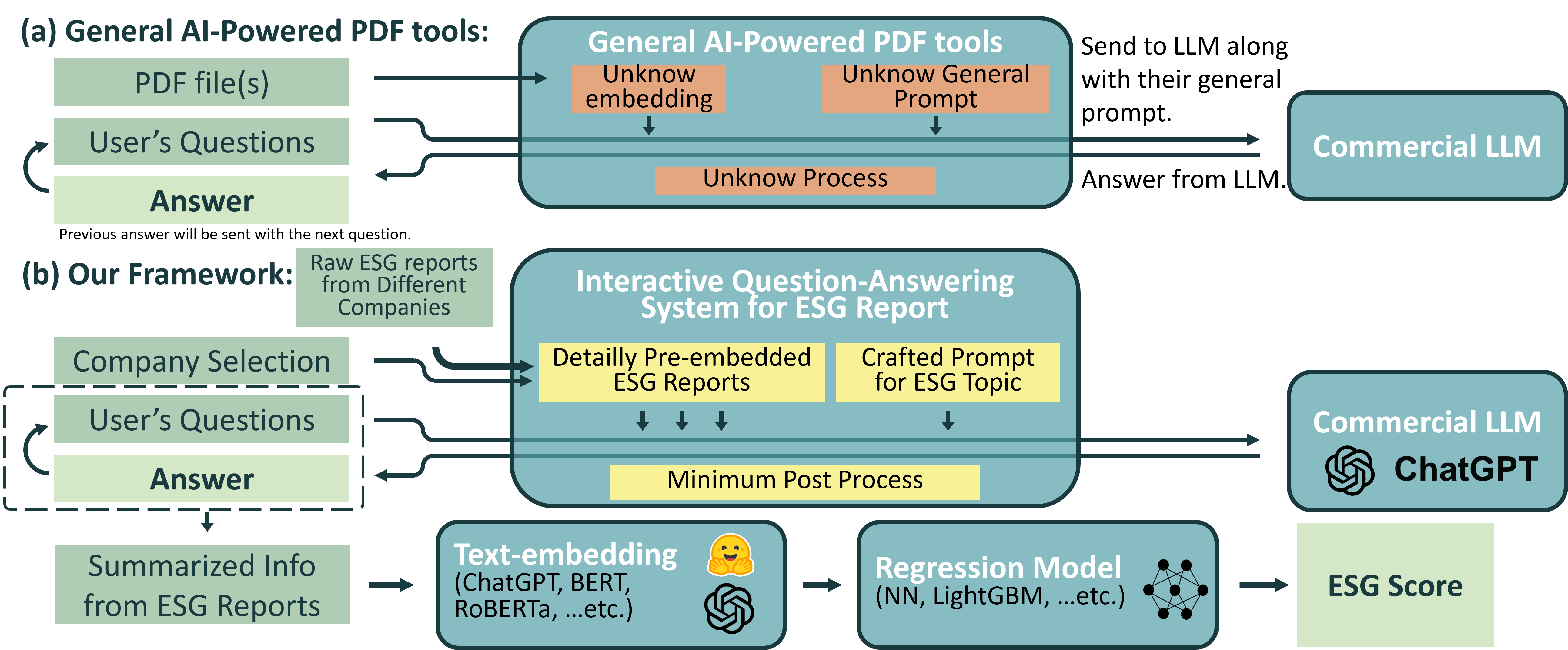}
\caption{Comparison between General AI-powered PDF tools and the proposed Interactive Question-Answering System for ESG Reports. (a) General AI-powered PDF tools typically take user-uploaded PDF files, process them with an unknown embedding model (primarily focused on text with limited support for equations, tables, charts, and images), and submit them to a commercial LLM using a generic prompt along with the user's questions, producing answers with unclear processing steps. (b) Our framework begins with the user selecting specific companies from a database to compare their ESG reports. The user then submits questions about one or more companies' ESG data. A crafted prompt tailored specifically to ESG topics is generated and sent to a commercial LLM (ChatGPT) via an API, minimizing post-processing. The resulting answer is delivered directly to the user interface.}
\label{NLP_1-1_diagram comparison}
\noindent\hrulefill
\end{figure*}

We present ESGLens as a proof-of-concept framework rather than a production system. Our primary contribution is the end-to-end pipeline design — from heterogeneous PDF processing through prompt-engineered extraction to regression-based ESG score prediction — validated on a limited dataset as a minimum viable prototype. We release the code to facilitate reproducibility and future extension.

\section{Related Work}

\subsection{LLM-Based ESG Report Analysis}
At the time of this project (late 2024 - early 2025), several concurrent efforts had begun applying LLMs to automate the analysis of corporate ESG disclosures. Ni et al.~\cite{ni2023chatreport} introduced ChatReport, an LLM-based tool for analyzing sustainability reports against TCFD guidelines, incorporating domain expert feedback to mitigate hallucinations. Zou et al.~\cite{zou2023esgreveal} applied LLMs combined with Retrieval-Augmented Generation (RAG) techniques~\cite{lewis2020retrieval}, building on the retrieval-augmented pre-training paradigm of REALM~\cite{guu2020realm}, to extract structured ESG data from corporate reports listed on the Hong Kong Stock Exchange, achieving 76.9\% data extraction accuracy with GPT-4.

Subsequent work (2025--2026) has further advanced this direction. Ding et al.~\cite{ding2025euleresgautomatingesgdisclosure} presented EulerESG, which combines dual-channel retrieval with SASB-aware metric extraction across multiple industries, achieving up to 0.95 average accuracy on standard-aligned metric tables. Hoang et al.~\cite{hoang2026esgreportinglifecyclemanagement} proposed an agentic framework covering the full ESG reporting lifecycle. Wu et al.~\cite{wu2024susgengptdatacentricllmfinancial} developed SusGen-GPT, a fine-tuned 7--8B parameter model that approaches GPT-4 performance on financial NLP and TCFD-compliant report generation tasks. Mousavian Anaraki et al.~\cite{mousavian-anaraki-etal-2025-automatic} addressed automatic GRI--SDG annotation with LLM-based filtering, while a companion survey~\cite{anaraki2025large} provided a systematic review of LLMs for sustainability reporting. ESGLens provides an alternative perspective by integrating GRI-guided extraction with quantitative score prediction via regression on LLM-generated embeddings -- a capability not addressed by these concurrent systems.

\subsection{ESG Text Classification and Domain-Specific Models}
Transformer-based language models~\cite{vaswani2017attention} have proven effective for ESG-related text classification. Pontes et al.~\cite{pontes2023leveraging} demonstrated the effectiveness of RoBERTa~\cite{liu2019roberta} and SVM for multilingual ESG issue classification. Tseng~\cite{tseng2023dynamicesg} proposed a dynamic approach to extracting ESG ratings from news articles in real time against MSCI and SASB standards. Xia~\cite{xia-etal-2024-using} combined PLMs and LLMs to develop ESGLlama and FinLlama for ESG and financial text classification, addressing the scarcity of domain-specific training data. Mehra et al.~\cite{mehra2022esgbert} developed ESGBERT, a language model tailored for ESG classification. However, these approaches primarily target news-level or sentence-level classification rather than full-document structured extraction from ESG reports.

\subsection{Positioning Relative to Generic PDF Tools}
While the above systems are purpose-built for ESG, many general-purpose AI-powered PDF tools (PDF.ai, ChatPDF, ChatDOC) leverage commercial LLMs to answer generic questions on arbitrary reports, as illustrated in Figure~\ref{NLP_1-1_diagram comparison}(a)~\cite{chatpdf2023, chatdoc2023, pdfai2024}. However, these tools lack domain-specific extraction guided by standards such as GRI, quantitative score prediction, and transparent RAG pipelines~\cite{mousavian-anaraki-etal-2025-automatic, hoang2026esgreportinglifecyclemanagement}. The precise methods they use to process PDFs and interface with LLMs remain opaque. ESGLens addresses these gaps through a domain-specific pipeline combining GRI-guided retrieval, prompt-engineered extraction, and regression-based scoring.

\section{Improved ChatReport: ESGLens}
In this work, we aim to gather the advantages of previous works and improve the efficiency of ESG report analysis. Contrary to generic PDF tools, ESGLens is architected around three core principles: (1) structured document processing tailored to heterogeneous ESG report formats, (2) GRI-standard guided extraction for standardized, comparable outputs, and (3) quantitative score prediction rather than mere summarization. ESG reports contain diverse content types including text, equations, tables, charts, images, and even chemical structures. For LLMs to fully comprehend these documents, all content should be converted into machine-readable syntax. Moreover, the extent of document embedding is field-specific; for instance, critical data in ESG reports are often found in tables and charts. To address these challenges and enhance the analysis of ESG reports, we propose ESGLens, a RAG-based framework illustrated in Figure \ref{NLP_1-1_diagram comparison}(b). This approach uses a RAG pipeline: embedding ESG reports from companies annually and storing them in a vector database, then retrieving context for both user queries and standardized GRI questions. The retrieved context is passed to the LLM via carefully engineered prompts, ensuring domain-specific extraction that ultimately feeds into a quantitative scoring component.

\section{Evaluation and Success Criteria}
\begin{itemize}[left=0cm, topsep=0.2cm, after=\vspace{-0.1cm}]
\item \textbf{Report Summary and Key Value Identification}: Initially, we manually identify key numbers such as GHG emissions from a selected ESG report. The next step involves testing the LLM's ability to accurately detect and extract these numbers.
\item \textbf{Hallucination Verification}: When extracting information from ESG reports, the model must also provide corresponding page numbers. This allows us to trace the extracted data back to the report, ensuring that the output is accurate and not a hallucination.
\item \textbf{Comparative Rating Analysis}: If the LLM is used to rate ESG reports, its rating results are compared with the official scores provided by LSEG to evaluate consistency and accuracy.
\end{itemize}

\section{Pertinent Dataset}
\begin{itemize}[left=0cm, topsep=0.2cm, after=\vspace{-0.1cm}]
\item \textbf{Responsibility Report}: A website for downloading ESG reports.
\item \textbf{LSEG Dataset}: A platform providing access to companies’ ESG scores and financial reports.
\item \textbf{DynamicESG}: A dataset for classifying ESG-related news.
\end{itemize}

\section{Implementation Details}
\begin{figure}[ht]
\includegraphics[width=15 cm]{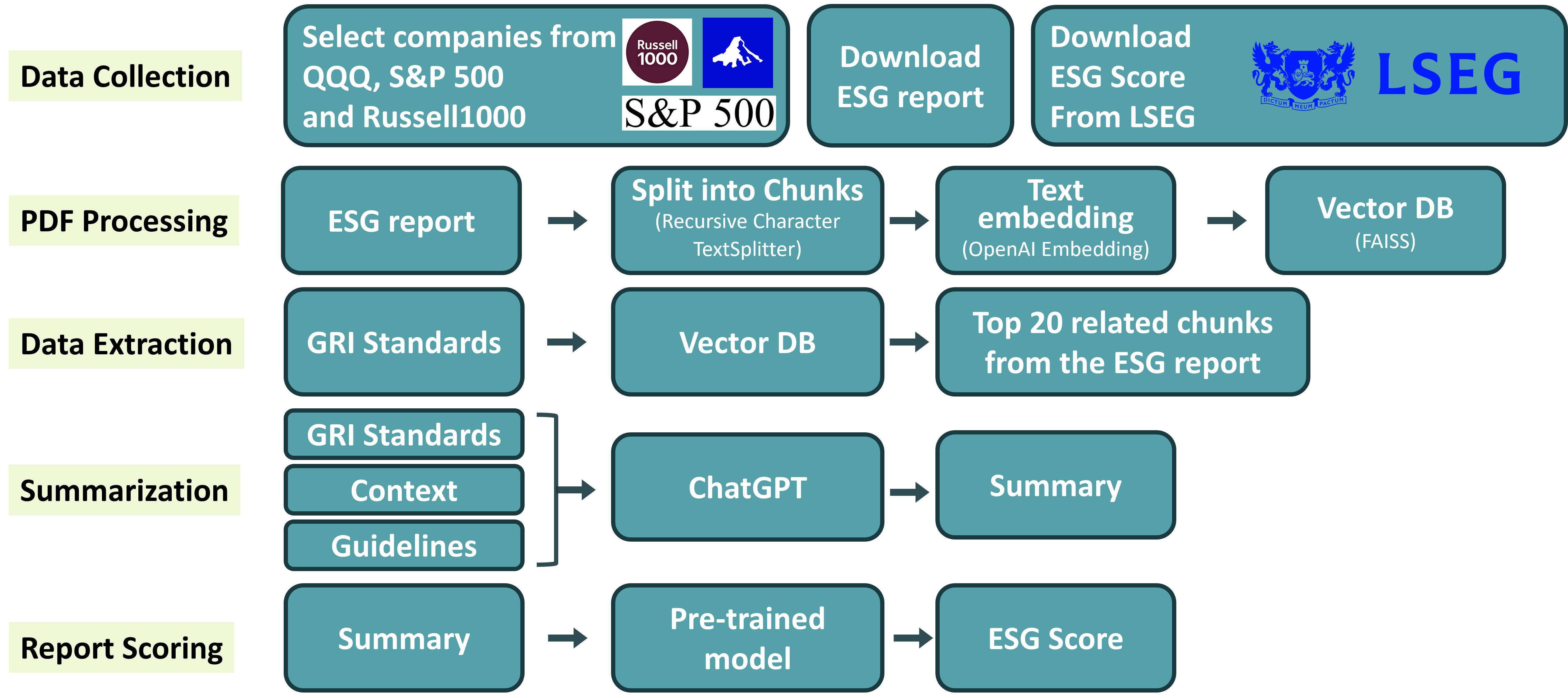}
\caption{Detailed process framework of ESGLens, illustrating the five-stage pipeline. 
(1)~\textbf{Data Collection}: selecting companies from the QQQ, S\&P~500, and Russell~1000 indices 
and downloading their ESG reports alongside LSEG reference scores; 
(2)~\textbf{PDF Processing}: splitting reports into chunks via RecursiveCharacterTextSplitter, 
embedding them with OpenAI embeddings, and storing them in a FAISS vector database; 
(3)~\textbf{Data Extraction}: -querying the vector database with GRI standards to retrieve 
the top~20 relevant chunks per standard; 
(4)~\textbf{Summarization}: combining GRI standards, retrieved context, and prompt guidelines 
as input to ChatGPT to produce structured summaries; and 
(5)~\textbf{Report Scoring}: embedding the summaries and feeding them into a regression model 
(Neural Network or LightGBM) to predict a quantitative ESG score, 
which is then benchmarked against LSEG ground-truth scores.}
\label{NLP_1-2_detailed process}
\end{figure}

ESGLens is divided into distinct modules: report processing, data extraction, and scoring, as shown in Figure \ref{NLP_1-2_detailed process}. To implement this, we use LangChain \cite{chase2022langchain}, a framework designed to facilitate the development of applications that integrate LLMs with other components. This framework was referenced from the Chatreport paper, which analyzes sustainability reports based on TCFD guidelines. However, we developed new prompts and workflows tailored to the GRI guidelines.

\begin{itemize}[left=0cm]
\item \textbf{PDF Processing (RAG Foundation)}: The first step in our RAG pipeline involves dividing the structured ESG report PDF into chunks. We use ChatGPT embeddings to embed these chunks into vector space using the RecursiveCharacterTextSplitter method. FAISS \cite{johnson2019billion} is employed for efficient vector database management and retrieval. This enables the retrieval component of our RAG architecture.

\item \textbf{GRI-Guided Data Extraction}: The GRI standards \cite{gri2021standards} contain approximately 120 guidelines. For this project, we start with five environment-related guidelines, as they are the most pressing topics. The module searches the vector database for the most relevant information related to these GRI standards, selecting the top 20 relevant chunks for each standard. Unlike generic question-answering, this step is explicitly guided by GRI reporting standards, ensuring domain-specific extraction.

\item \textbf{Prompt-Engineered GRI Extraction}: Once the top 20 relevant chunks are identified via RAG, we design prompts that include CONTEXT, QUESTIONS, and GUIDELINES. The CONTEXT is based on the retrieved chunks, and the QUESTIONS pertain to specific GRI standards. The GUIDELINES ensure that ChatGPT provides accurate, standards-compliant responses—moving beyond generic summarization.

\item \textbf{Quantitative Score Prediction}: After extracting the relevant information from the report, we compile bullet points related to each GRI standard. These structured extractions are embedded and fed into a regression model to predict the company's ESG score. This quantitative output (not just summaries) is cross-referenced with official LSEG scores. The regression model can be trained with various embeddings (ChatGPT, BERT \cite{devlin2019bert}, RoBERTa) and model types (Neural Networks, LightGBM \cite{ke2017lightgbm}).
\end{itemize}

\section{Results and Analysis}\label{Preliminary}
\subsection{Prompt Engineering}
To effectively extract information from a long document and to ensure that the LLM output follows a specific format, it is essential to craft prompts that accurately elicit the desired information. A prompt to the LLM usually consists of the following parts:
\begin{itemize}[left=0cm]
\item \textbf{Instruction}: An opening higher-level description about the task or a guidance that ask model to play a certain role. In this application, the opening usually starts with "You are tasked with the role of an ESG expert, assigned to analyze a company's ESG report." This part might also include example about how to answer the question.
\item \textbf{Context}: The document that we want the model to look into. It our case, the context would be the chunk of ESG report that relates to the question.
\item \textbf{Question}: The specific query directed at the provided context. For instance, one of the questions in our case would be "GRI\_302\_1: What is the total energy consumption within the organization, including all forms of energy used?".
\end{itemize}

The position of the question within the prompt was found to influence the quality of the extracted information. Furthermore, utilizing few-shot learning by providing examples of the desired output structure can improve output clarity. Three different prompt strategies and their results are presented in section \ref{sec_appendix}. In Prompt 1, where the question appears at the end of the prompt, only one result was generated. In Prompt 2, where the question directly follows the context, more comprehensive information was extracted, but the lack of a specified output format led to unstructured results. Prompt 3 includes examples that illustrate how content and sources should be organized, enabling the LLM to generate a clean, structured summary.

\begin{table}[ht]
\centering
\caption{Comparison of prompt structure and results.}
\label{t_prompt_structure}
\renewcommand{\arraystretch}{1.4}
\begin{tabular}{c p{7cm} p{7cm}}
\toprule
\textbf{Prompt} & \textbf{Structure} & \textbf{Result} \\
\midrule
1 & Instruction, question, context, and a final instruction. & Fair. Only one result was generated. \\
2 & Instruction, context, and question. & Good. More comprehensive results, but lacked a specified output format. \\
3 & Instruction with examples, context, and question. & Best. Generated a clean and structured summary. \\
\bottomrule
\end{tabular}
\end{table}

\subsection{Traceability}
To ensure that ESGLens extracts accurate information from the document, the temperature setting was set to 0 to minimize hallucination. Additionally, the source of each extracted piece of information was required to be included with the output, enabling traceability. This approach allows verification of the accuracy of ESGLens's responses by cross-referencing them with the original document. As shown in Table \ref{t_ext_info}, we asked the model one question: "What is the total energy consumption within the organization, including all forms of energy used?" The model extracted 10 items related to energy consumption from the document, of which 8 items were found in the original document and accurately identified, demonstrating strong retrieval capabilities.

The two failures merit detailed analysis. Item 2 reflects a \emph{prompt scope} issue: the model did not distinguish between planned actions and those already implemented, owing to the prompt's focus on energy consumption without temporal context. Item 10 reveals a \emph{prompt precision} issue: the initial prompt designed to locate energy consumption mistakenly included water consumption data. Thus, developing a more refined prompt is crucial. Items 3 and 4, however, represent a distinct failure mode: \emph{few-shot example leakage}. Both claims are near-verbatim copies of the demonstration examples provided in Prompt~3 to illustrate the desired JSON output format (see Appendix), with only the source page numbers altered. The model treated illustrative content as factual extractions rather than as structural templates. This risk is amplified when few-shot examples are topically similar to the query. A practical mitigation is to use topically dissimilar examples (such as demonstrating the output format with governance or social metrics when querying environmental data), so that the model cannot confuse template content with retrieved context.

\subsection{LSEG scoring}
LSEG quantifies a company's ESG performance using two types of data: (1) Boolean and (2) numeric data. \cite{lseg2024esg} Boolean data points are assigned values of 1 for 'Yes' and 0 for 'No' or 'Null,' with a default value of 0 applied when relevant data is unavailable. Each measure has a polarity to indicate whether higher values are positive or negative, and these are converted to numeric values for percentile score calculations. ESG scores are derived from 10 weighted categories, grouped under the three ESG pillars: Environment (Emissions, Innovation, Resource Use), Social (Human Rights, Workforce, Community), and Governance (Management, Shareholders, CSR Strategy). The category weights are calculated by summing up magnitude weights for each industry group and normalizing them to reflect the relative importance of each category within the group.

\begin{figure}[H]
\centering
\includegraphics[width=15 cm]{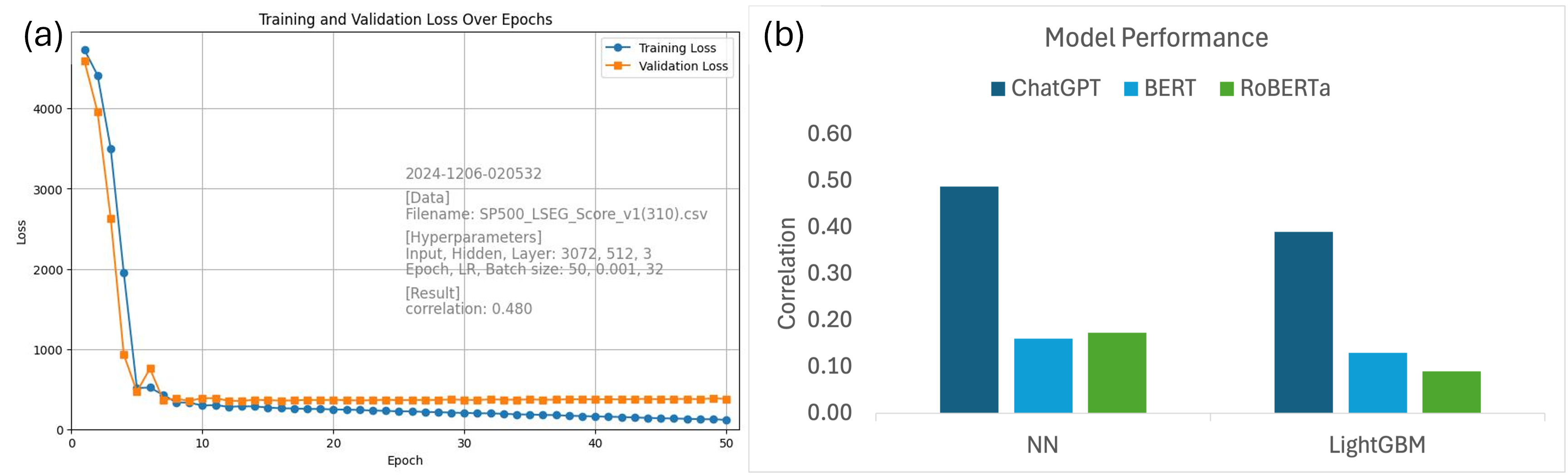}
\caption{(a) Training loss of NN. (b) Correlation of predicted and actual ESG scores using ChatGPT, BERT, and RoBERTa embeddings with Neural Network and LightGBM models. ChatGPT outperforms the others.}
\label{NLP_2-1_trmp}
\end{figure}

\subsection{ESG scoring prediction}
To emulate LSEG’s ESG evaluation approach, we developed a prompt designed to analyze ESG reports. This prompt includes questions such as identifying the company’s industry group and its emissions for a specific year. Additionally, we required the model to evaluate the tone (positive or negative) and extract numeric values (e.g. 810 tons of CO\textsubscript{2} emission) from the provided context to supply data for our scoring model. Examples of categorized ESG information with tone and value assessments are shown in Table~\ref{t_ext_cate}. The complete list of questions is provided in the Appendix (see Table~\ref{tab:environment_questions}), along with the detailed prompt used for this evaluation. 

To predict the ESG score of a report based on its summary, we first transformed the summary into embeddings and then input these embeddings into a regression model to generate predictions. The regression model was trained using the ESG scores provided by LSEG. We evaluated three embedding methods – ChatGPT, BERT, and RoBERTa – and two regression models : a Neural Network and LightGBM. \cite{lee2022esg}
The Neural Network architecture consisted of three layers, with an input size of 3072 (matching the embedding dimensions) and 512 neurons in each hidden layer. The model was trained for 50 epochs with a learning rate of 0.001 and a batch size of 32. The training loss exhibited rapid convergence, stabilizing within the first 10 epochs, as illustrated in Figure \ref{NLP_2-1_trmp}(a). For the LightGBM model, the hyperparameters were configured as follows: max bin of 512, learning rate of 0.01, 50 iterations, minimum gain to split of 0.01, and a feature fraction of 0.8.

The results, as depicted in Figure \ref{NLP_2-1_trmp}(b), show that ChatGPT embeddings with the Neural Network regressor achieve the highest Pearson correlation of $r \approx 0.48$ ($R^{2} \approx 0.23$), meaning the model captures roughly 23\% of the variance in LSEG ground-truth scores. While modest in absolute terms, this result should be interpreted in light of two factors: (1)~the training set comprises only ${\sim}300$ reports, which limits the capacity of any supervised regressor, and (2)~the prediction is derived entirely from LLM-extracted \emph{text} summaries of the environmental pillar alone, without access to the Boolean and numeric data points that LSEG uses across all three pillars. As a reference baseline, a mean-prediction model (predicting the training-set average for every report) yields $r = 0$ by construction; BERT and RoBERTa embeddings under the same Neural Network architecture reach only $r \approx 0.16$ and $r \approx 0.17$, respectively, confirming that the observed signal is embedding-dependent rather than an artifact of the regression head. LightGBM trained on ChatGPT embeddings yields $r \approx 0.39$, suggesting that the Neural Network's additional capacity is beneficial at this scale. A detailed summary of emissions goals and achievements across the analyzed reports is provided in Table~\ref{tab:emissions_summary}.

Items~3 and~4, however, represent a distinct failure mode: \emph{few-shot example leakage}.  Both claims are near-verbatim copies of the demonstration examples provided in Prompt~3 to illustrate the desired JSON output format (see Appendix), with only the source page numbers altered. The model treated the illustrative content as factual extractions rather than as a structural template. This risk is amplified when the few-shot examples are topically similar to the query, as was the case here (both concerned energy consumption). A practical mitigation is to use topically dissimilar examples (like demonstrating the output format with a governance or social metric when querying environmental data), therefore the model cannot confuse template content with retrieved context.

\section{Conclusion}
We presented ESGLens, a novel framework that distinguishes itself from generic PDF tools through four key design principles: (1) purpose-built processing for heterogeneous structured ESG documents, (2) GRI-standard guided extraction rather than generic question-answering, (3) quantitative score prediction as the primary output, and (4) a transparent RAG+embedding pipeline enabling reproducibility. 

We collected ESG reports and corresponding LSEG scores for companies listed in the QQQ, S\&P 500, and Russell 1000 indices as of 2022. ESGLens employs a systematic RAG process, starting with the segmentation of documents into chunks, embedding each via LLM in vector space, and storing in a vector database. Relevant chunks are retrieved in response to both user queries and predefined GRI-aligned questions. This retrieved context is integrated with tailored prompts to generate GRI-compliant extractions that ultimately feed into quantitative score predictions. 

The GRI-guided extracted information is then processed through our quantitative scoring component, where it is embedded and input into a regression model to predict the ESG score for a company. This quantitative output represents a key differentiator from generic summarization tools. Examples of environmental data extraction across multiple categories are shown in Table~\ref{tab:environmental_summary}. Our experiments evaluated various prompts and embedding methods, including ChatGPT, BERT, and RoBERTa. Among these, ChatGPT embeddings outperformed other models. 

Furthermore, the comparison of regression models revealed that the Neural Network model outperformed LightGBM, achieving a Pearson correlation of $r \approx 0.48$ ($R^{2} \approx 0.23$) between the predicted and actual ESG scores. This level of explained variance is consistent with the constrained training set (${\sim}300$ reports) and the restriction to environmental indicators only; we expect meaningful gains from expanding both the dataset and pillar coverage. Mean absolute error (MAE) and root mean squared error (RMSE) are deferred to future work with larger training sets where they would provide more informative comparative signals. These findings demonstrate that even a text-only, single-pillar pipeline can recover a statistically meaningful signal from ESG reports, supporting the viability of ESGLens's domain-specific RAG+LLM architecture as a foundation for more comprehensive automated ESG analysis.

\subsection{Limitations}
This framework can be further improved in several ways:
\begin{itemize}[left=0cm]
\item  \textbf{Dataset Size}: The current model's accuracy is constrained by the small dataset. Many ESG reports from 2022 were either incomplete or varied significantly in structure, rendering some reports unprocessable. Additionally, not all company ESG scores were available in the LSEG database, reducing the training set to approximately 300 data points. Expanding the dataset across fiscal years and addressing structural heterogeneity could yield meaningful performance gains.
\item  \textbf{Expanding to Social and Governance Pillars}: The analysis currently focuses solely on the environmental component. The social and governance dimensions also contribute significantly to the overall ESG score and should be incorporated to provide a holistic assessment.
\item  \textbf{Multimodal Content Extraction}: Most ESG reports include detailed tables, charts, and images containing critical quantitative information. The current pipeline segments documents into text chunks without dedicated handling of non-textual elements. Integrating table extraction and chart understanding modules could substantially improve quantitative accuracy.
\end{itemize}

\subsection{Future Work}
Beyond addressing the limitations above, several directions merit exploration. First, the emergence of dedicated ESG benchmarks, like ESGenius~\cite{He_2025}, which evaluates 50~LLMs on ESG question answering across multiple standards, and MMESGBench~\cite{zhang2025mmesgbenchpioneeringmultimodalunderstanding}, which targets multimodal reasoning over long-form ESG documents. They provides an opportunity to rigorously evaluate ESGLens's extraction and scoring components against standardized baselines. At the time of this project, no such benchmarks existed; incorporating them in future iterations would strengthen the empirical grounding of the framework.

Second, the recent proliferation of agentic and lifecycle-oriented ESG systems~\cite{hoang2026esgreportinglifecyclemanagement, ding2025euleresgautomatingesgdisclosure} suggests that ESGLens's modular architecture could be extended with autonomous agent workflows. For example, automatically detecting newly published reports, triggering extraction, and updating score predictions in a continuous monitoring loop. Third, replacing the current text-only embedding pipeline with multimodal document representations that jointly encode tables, charts, and text could address the most significant information loss in the current design.

\section{Acknowledgement}
We thank Sara Lin (CPA) for approaching us with this research idea.

\section{Code and Data Availability}
The code for ESGLens is available at \url{https://github.com/tyy4ng/ESGLens}, 
which includes modules for LSEG score collection and ESG score prediction.

\clearpage
\onecolumn
\section*{Appendix}\label{sec_appendix}
\subsection*{Prompts}
\begin{mdframed}[frametitle=Prompt 1]
You are tasked with the role of an ESG expert, assigned to analyze a company's ESG report. Based on the following extracted parts from the sustainability report, answer the given QUESTIONS. If you don't know the answer, just say that you don't know. Don't try to make up an answer.

\vspace{0.3cm}

Format your answers in JSON format and list up to 10 related points. Each point should include its corresponding SOURCE.

\vspace{0.3cm}

\textbf{QUESTIONS:} \\
\hspace{1em} GRI\_302\_1: What is the total energy consumption within the organization, including all forms of energy used? 

\vspace{0.3cm}

\texttt{=========} \\
\texttt{\{context\}} \\
\texttt{=========}

\vspace{0.3cm}

\# Your FINAL\_ANSWER in JSON (ensure there's no format error).
\end{mdframed}

\begin{mdframed}[frametitle=Result 1]
\label{lst:result1_gri302_1}
\begin{lstlisting}[language=json]
{
    "GRI_302_1": {
        "Total_Energy_Consumption": "810 GWh", 
        "Source": "93"
    }
}
\end{lstlisting}
\end{mdframed}
\vspace{0.5cm}

\begin{mdframed}[frametitle=Prompt 2]
You are tasked with the role of an ESG expert, assigned to analyze a company's ESG report. 
Based on the following extracted parts from the sustainability report, answer the given QUESTIONS. 
If you don't know the answer, just say that you don't know. Don't try to make up an answer.

\vspace{0.3cm}
Format your answers in JSON format and list up to 10 related points. Each point should include its corresponding SOURCE.

\vspace{0.3cm}
\textcolor{blue}{Now, answer the following question using the provided \{context\}:}

\vspace{0.3cm}
\textcolor{blue}{GRI\_302\_1: What is the total energy consumption within the organization, including all forms of energy used?}
\end{mdframed}

\begin{mdframed}[frametitle=Result 2-1]
\label{lst:result2_1_energy_metrics}
\begin{lstlisting}[language=json]
total energy consumption: { 
value: 280 GWh annually, 810 GWh accumulative, 
source: 93 
},
energy efficiency target failure: { 
description: The energy efficiency of 5nm volume production failed to achieve the annual target. The Company plans to expand energy-saving actions and invest resources to improve energy efficiency., 
source: 105
},
energy saving measures: { 
measures: [
  Smart lighting in non-cleanroom areas, 
  Replace bulbs with LED lighting, 
  AI powered chilled water energy saving system, 
  Replace cold water with PCW for cooling air compressors throughout first- and second-stage compression, 
  Modify wet film plate for makeup air handling unit (MAU),                                     
  Energy-efficient fan blades for PCW towers,                                                           
  Replace tool components with energy-saving components,                                                         
  Hot DI water circulation system
], 
source: 107 
},
\end{lstlisting}
\end{mdframed}

\begin{mdframed}[frametitle=Result 2-2]
\label{lst:result2_2_conservation_results}

\begin{lstlisting}[language=json]
energy conservation_results: { 
annual total energy reduction: 280 GWh, 		       
accumulative total energy reduction: 810 GWh, 
annual total water reduction: 13.5 million metric tons, 
accumulative total water reduction: 42.58 million metric tons, 
waste production reduction among major waste-producing suppliers: 39%, 
ISO 14064-1 verification for 84% of high-energy consumption suppliers: Verified, 
requirement for suppliers to introduce energy-saving evaluation when building new plants: Required, 
source: 93
}
\end{lstlisting}
\end{mdframed}
\vspace{0.5cm}

\begin{mdframed}[frametitle=Prompt 3]
    You are tasked with the role of an ESG expert, assigned to analyze a company's ESG report. 
    Based on the following extracted parts from the sustainability report, answer the given QUESTIONS. 
    If you don't know the answer, just say that you don't know. Don't try to make up an answer.

    \vspace{0.3cm}

    Format your answers in JSON format and list up to 10 related points. Each point should include its corresponding SOURCE.

    \vspace{0.3cm}

    \textcolor{blue}{Your FINAL\_ANSWER should be structured as follows:}\\
    
    \begin{lstlisting}[language=json, basicstyle=\ttfamily\color{blue}]
    \label{lst:prompt3_example_output}
    {
      "Summary": [
        {
          "point": "Energy consumption in 2023 was 810 GW, including electricity, natural gas, and renewable sources.",
          "source": "12"
        },
        {
          "point": "The company reported a 5% increase in renewable energy use, reaching 20% of total energy consumption.",
          "source": "14"
        },
        {
          "point": "The total energy consumption for electricity alone was 500 GW.",
          "source": "16"
        }
      ]
    }
    \end{lstlisting}
    \vspace{0.3cm}

    Now, answer the following question using the provided \{context\}:
    
    \vspace{0.3cm}
    
    GRI\_302\_1: What is the total energy consumption within the organization, including all forms of energy used?
\end{mdframed}

\begin{mdframed}[frametitle=Result 3-1]
\label{lst:result3_1_energy_reduction}
\begin{lstlisting}[language=json]
{
    "Summary": [
        {
            "point": "The annual total energy reduction reached 280 GWh, and the accumulative total reached 810 GWh.",
            "source": "93"
        },
        {
            "point": "The company implemented 822 energy-conservation programs, resulting in a total reduction of 800 GWh in electricity consumption.",
            "source": "20"
        },
        {
            "point": "The company reported a 5% increase in renewable energy use, reaching 20% of total energy consumption.",
            "source": "4"
        },
        {
            "point": "The total energy consumption for electricity alone was 500 GW.",
            "source": "5"
        },
        {
            "point": "The company planned and carried out 822 energy-saving measures across eight major categories, saving an additional 830 GWh.",
            "source": "100"
        },
    ]
}
\end{lstlisting}
\end{mdframed}

\begin{mdframed}[frametitle=Result 3-2]
\label{lst:result3_2_achievements}
\begin{lstlisting}[language=json]
{
    "Summary": [
        {
            "point": "The company missed the annual target for the energy efficiency of 5nm volume production.",
            "source": "105"
        },
        {
            "point": "The company aims to increase the use of renewable energy to 60% by 2030.",
            "source": "20"
        },
        {
            "point": "The company requires suppliers to introduce energy-saving evaluation when building new plants.",
            "source": "93"
        },
        {
            "point": "The company reduced supplier energy consumption by a cumulative total of 810 GWh, surpassing the target of 550 GWh.",
            "source": "78"
        },
        {
            "point": "The company reduced supplier water consumption by a cumulative total of 42.58 million metric tons, surpassing the target of 30 million metric tons.",
            "source": "78"
        }
    ]
}
\end{lstlisting}
\end{mdframed}

\vspace{0.5cm}
\begin{mdframed}[frametitle=Questions]
\label{lst:esg_evaluation_questions}
\begin{lstlisting}[language=json]
{'emissions_1': 'What are the greenhouse gas emissions for Scope 1, Scope 2, and Scope 3?',
'emissions_2': 'What is the carbon intensity (emissions per unit of product or revenue)?',
'emissions_3': 'What are the carbon reduction targets, and what progress has been made?',
'emissions_4': 'How are air pollutants, such as nitrogen oxides and sulfur oxides, managed?',
'resource_1': 'What is the total energy consumption, and what proportion of it comes from renewable sources?',
'resource_2': 'How is water used, including total water consumption and the recycling rate?',
'resource_3': 'How efficient is the use of raw materials, and what recycling initiatives are in place?',
'resource_4': 'How is waste managed, and what are the methods and recycling targets?',
'environment_1': 'What green products or services does the company offer?',
'environment_2': 'How much investment has been made in environmental innovation or R&D? What projects are included?',
'environment_3': 'Is the company participating in initiatives promoting the circular economy?',
'biodiversity_1': 'What policies are in place to protect biodiversity or natural habitats?',
'biodiversity_2': 'Does the company participate in biodiversity conservation projects? If so, what are they?',
'biodiversity_3': 'What is the impact of the supply chain on land use and the environment?',
'ems_1': 'Does the company have publicly available environmental policies? If so, what are they?',
'ems_2': 'Has the company obtained certifications such as ISO 14001 or similar? If so, which ones?',
'ems_3': 'How transparent is the company regarding environmental data and goals?'}
\end{lstlisting}
\end{mdframed}

\begin{mdframed}[frametitle=Prompt 4]
    You are an ESG (Environmental, Social, and Governance) expert assigned to analyze a company's sustainability report. Your task is to analyze the extracted parts of the report and answer the given QUESTIONS. If the information required to answer a question is not present in the provided context, respond with: "No data available." Do not fabricate answers.
    
    Your response must follow these requirements:
    1. **Format your answer in JSON** with three structured sections: `ContentExtraction`, `ToneAnalysis`, and `NumericalData`.
    2. Each section must list **exactly one point**, including its corresponding `source`, `tone`, and (if applicable) `value`.
    3. **Tone** must be classified as one of: `Neutral`, `Positive`, or `Negative`.
    4. **Source** must be an integer and represent the associated reference number for the information.
    
    \vspace{0.3cm}

    \textcolor{blue}{Your FINAL\_ANSWER should be structured as follows:}\\
    
    \begin{lstlisting}[language=json, basicstyle=\ttfamily\color{blue}]
    \label{lst:prompt4_example_output}
    Your FINAL_ANSWER should be structured as follows:
    {{
        "ContentExtraction": [
            {{
                "point": "The company aims to reduce carbon emissions by 30% by 2025.",
                "tone": "Positive",
                "value": "Null",
                "source": 8
            }}
        ]
    }}
    
    ### **Handling Missing Data**
    If relevant data is not available for a specific section, use the following structure:
    {{
        "ContentExtraction": [
            {{
                "point": "No data available.",
                "tone": "Neutral",
                "value": "Null",
                "source": 0
            }}
        ]
        ]
    }}
    \end{lstlisting}
    \vspace{0.3cm}
    **Instructions**
    Now, using the provided \{context\} and \{question\}, answer the question while adhering to the JSON format above. Ensure each section is labeled and includes all required fields (`point`, `tone`, `value`, `source`).
    \vspace{0.3cm}
    
\end{mdframed}

\newpage

\subsection*{Supplementary Tables}
\nopagebreak[4]
\begin{table}[H]
\centering
\caption{Comparison of extracted information and documented statements with correctness verification.}
\label{t_ext_info}
\renewcommand{\arraystretch}{1.2}
{\small 
\begin{tabular}{c p{4.5cm} p{9cm} p{1.5cm}}
\toprule
\textbf{Item} & \textbf{Extracted Information} & \textbf{Paragraph in Document} & \textbf{Correctness} \\ \midrule
1 & The annual total energy reduction reached 280 GWh, and the cumulative total reached 810 GWh. & ... obtained an average grade of B-, exceeding the initial target of C. TSMC requires and assists suppliers to improve their sustainability performance. In 2023, the annual total energy reduction reached 280 GWh, and the cumulative total reached 810 GWh. The annual total water reduction reached 13.5 million metric tons, and the cumulative total reached 42.58 million metric tons. Additionally, the waste production among major ... & \checkmark \\ \midrule
2 & The company implemented 822 energy-conservation programs, resulting in a total reduction of 800 GWh in electricity consumption. & ... Made significant strides in climate change mitigation, achieving net-zero emissions in scope 1 and scope 2 at TSMC overseas operations; expanded its onshore wind farm power supply, increasing the proportion of renewable energy usage from 40\% to 60\% by 2030, and proposed 822 energy-conservation programs, resulting in a total reduction of 800 GWh in electricity consumption. ... & \checkmark \\ \midrule
3 & The company reported a 5\% increase in renewable energy use, reaching 20\% of total energy consumption. & (No related information in the ESG report.) & \xmark \\ \midrule
4 & The total energy consumption for electricity alone was 500 GW. & (No related information in the ESG report.) & \xmark \\ \midrule
5 & The company planned and carried out 822 energy-saving measures across eight major categories, saving an additional 830 GWh. & ... Planned and carried out 822 energy-saving measures across eight major categories, saving an additional 830 GWh. See Increase Energy Efficiency for more information. & \checkmark \\ \midrule
6 & The company missed the annual target for the energy efficiency of 5nm volume production. & ... In 2023, TSMC consumed a total of 24,700 GWh in energy, which resulted in a 0.1\% shortfall from its 5nm production efficiency target. This was mainly due to global supply chain disruptions. The failure to achieve this target prompted TSMC to invest further in energy-saving measures and implement additional guidelines. & \checkmark \\ \midrule
7 & The company targets increased use of renewable energy to 60\% by 2030. & ... Made significant strides in climate change mitigation, achieving net-zero emissions in scope 1 and scope 2 at TSMC overseas operations; expanded its onshore wind farm power supply, increasing the proportion of renewable energy usage from 40\% to 60\% by 2030, and proposed 822 energy-conservation programs, resulting in a total reduction of 800 GWh in electricity consumption. ... & \checkmark \\ \midrule
8 & The company requires suppliers to introduce energy-saving evaluation when building new plants. & ... target of C. TSMC requires and assists suppliers to improve their sustainability performance. In 2023, the annual total energy reduction reached 280 GWh, and the cumulative total reached 810 GWh. The annual total water reduction reached 13.5 million metric tons, ... & \checkmark \\ \midrule
9 & The company reduced supplier energy consumption by a cumulative total of 810 GWh, surpassing the target of 550 GWh. & $\uparrow$ Reduced supplier energy consumption by a cumulative total of 810 GWh. Target: 550 GWh. & \checkmark \\ \midrule
10 & The company reduced supplier water consumption by a cumulative total of 42.58 million metric tons, surpassing the target of 30 million metric tons. & $\uparrow$ Reduced supplier water consumption by a cumulative total of 42.58 million metric tons. Target: 30 million metric tons. & \checkmark \\
\bottomrule
\end{tabular}}
\end{table}

\onecolumn
\begin{table}[htbp]
\centering
\caption{Categorized summary of extracted ESG information with tone and value assessment.}
\label{t_ext_cate}
\renewcommand{\arraystretch}{1.2}
{\small 
\begin{tabular}{c p{9cm} p{2cm} c}
\toprule
\textbf{Category} & \textbf{Extracted Information} & \textbf{Tone} & \textbf{Value} \\ \midrule

emissions\_1 & Scope 1 (gross emissions) for fiscal year 2021 was 55,200 metric tons CO2e. & Neutral & 55,200 \\ \midrule
emissions\_1 & Scope 2 (market-based) emissions for fiscal year 2021 were 2,780 metric tons CO2e. & Neutral & 2,780 \\ \midrule
emissions\_1 & Scope 3 (gross emissions) for fiscal year 2021 were 23,130,000 metric tons CO2e. & Neutral & 23,130,000 \\ \midrule
emissions\_2 & The company targets a 30\% reduction in carbon emissions by 2025. & Positive & 30\% \\ \midrule
emissions\_3 & Manufacturing emissions account for approximately 70\% of the company's gross carbon footprint. & Positive & 70\% \\ \midrule
emissions\_4 & Manufacturing emissions account for approximately 70\% of the company's gross carbon footprint. & Negative & --- \\ \midrule
emissions\_4 & Several components require manufacturing processes using fluorinated gases with high global warming potential. & Negative & --- \\ \midrule
resource\_1 & Total energy consumption and proportion from renewable sources (2021): 18,100,000 MWh (100\% renewable); (2020): 11,400,000 MWh (100\% renewable). & Neutral & --- \\ \midrule
resource\_2 & Recycled water is utilized for irrigation, make-up water in cooling, and toilet flushing. & Neutral & --- \\ \midrule
resource\_3 & Transitioning all materials in products and packaging to recycled and renewable sources. & Positive & --- \\ \midrule
environment\_1 & The company has achieved carbon neutrality for operations since 2020 and targets carbon neutrality across the entire business by 2030. & Positive & --- \\ \midrule
environment\_2 & Scope 3 emissions calculations initiated in fiscal year 2017, including electricity transmission and distribution losses of approximately 28,000 metric tons CO2e. & Neutral & 28,000 \\ \midrule
environment\_3 & The company targets a 30\% reduction in carbon emissions by 2025. & Positive & 30\% \\ \midrule
biodiversity\_1 & The company targets a 30\% reduction in carbon emissions by 2025. & Positive & 30\% \\ \midrule
biodiversity\_2 & The company participates in biodiversity conservation projects, including mangrove conservation and ecosystem restoration initiatives. & Positive & --- \\ \midrule
biodiversity\_3 & The company targets carbon neutrality across its entire product footprint by 2030. & Positive & --- \\ \midrule
ems\_1 & The company targets carbon neutrality across its entire business by 2030. & Positive & --- \\ \midrule
ems\_2 & The company maintains ISO 27001 and ISO 27018 certifications for information security. & Positive & --- \\ \midrule
ems\_3 & The company targets carbon neutrality across the entire business, including the full product life cycle, by 2030. & Positive & ---\\ 
\bottomrule
\end{tabular}} 
\end{table}

\newpage

\begin{table}[!htbp]
\centering
\renewcommand{\arraystretch}{1.2}
\caption{Environmental Data Summary}
\label{tab:environmental_summary}
\begin{tabular}{l p{7cm} l r c}
\toprule
\textbf{Category} & \textbf{Summary} & \textbf{Tone} & \textbf{Value} & \textbf{Page} \\ 
\midrule
\texttt{emissions\_1} & Scope 1 (gross emissions) for fiscal year 2021 was 55,200 metric tons CO2e. & Neutral & 55,200 & 78 \\ 
\midrule
\texttt{emissions\_1} & Scope 2 (market-based) emissions for fiscal year 2021 were 2,780 metric tons CO2e. & Neutral & 2,780 & 78 \\ 
\midrule
\texttt{emissions\_1} & Scope 3 (gross emissions) for fiscal year 2021 were 23,130,000 metric tons CO2e. & Neutral & 23,130,000 & 78 \\ 
\midrule
\texttt{emissions\_4} & Manufacturing emissions account for approximately 70\% of the company's gross carbon footprint. & Negative & --- & 14 \\ 
\midrule
\texttt{emissions\_4} & Several components require manufacturing processes using fluorinated gases with high global warming potential. & Negative & --- & 16 \\ 
\midrule
\texttt{resource\_1} & Total energy consumption and proportion from renewable sources (2021): 18,100,000 MWh (100\% renewable); (2020): 11,400,000 MWh (100\% renewable); (2019): 5,700,000 MWh (100\% renewable); (2018): 4,100,000 MWh (99\% renewable); (2017): 1,900,000 MWh (97\% renewable). & Neutral & --- & 16 \\ 
\midrule
\texttt{resource\_2} & Recycled water is utilized for irrigation, make-up water in cooling, and toilet flushing. & Neutral & --- & 80 \\ 
\midrule
\texttt{resource\_3} & Transitioning all materials in products and packaging to recycled and renewable sources. & Positive & --- & 76 \\ 
\bottomrule
\end{tabular}
\end{table}

\begin{table}[!htbp]
\centering
\caption{Summary of emissions goals and reported achievements.}
\label{tab:emissions_summary}
\renewcommand{\arraystretch}{1.2}
\begin{tabular}{l p{9cm} l l c}
\toprule
\textbf{Category} & \textbf{Summary} & \textbf{Tone} & \textbf{Value} & \textbf{Page} \\
\midrule

\texttt{emissions\_1} & Compared to the previous year, overall Scope 3 emissions decreased by approximately 25\%, driven by reduced production volume in 2022, increased renewable energy usage, and updated EPA emissions factors. & Positive & --- & 25 \\

\texttt{emissions\_2} & The company targets a 55\% reduction in Scope 3 emissions per million USD of value added by 2030. & Positive & --- & 10 \\

\texttt{emissions\_3} & The company aims to reduce emissions by approximately 50\% by 2030 and achieve net-zero emissions thereafter. & Positive & --- & 10 \\

\texttt{emissions\_4} & The company targets a 30\% reduction in carbon emissions by 2025. & Positive & 30\% & 3 \\

\texttt{resource\_1} & The company aims to increase the share of onsite generation and externally supplied clean electricity. & Positive & --- & 19 \\

\texttt{resource\_2} & The company targets a 30\% reduction in carbon emissions by 2025. & Positive & 30\% & 3 \\

\texttt{resource\_3} & The company targets a 30\% reduction in carbon emissions by 2025. & Positive & 30\% & 3 \\

\bottomrule
\end{tabular}
\end{table}

\begin{table}[t]
\centering
\renewcommand{\arraystretch}{1.2}
\caption{Environmental Questions by Category}
\label{tab:environment_questions}
\begin{tabular}{l p{11.5cm}}
\toprule
\textbf{Category} & \textbf{Questions} \\ 
\midrule
\texttt{emissions\_1} & What are the greenhouse gas emissions for Scope 1, Scope 2, and Scope 3? \\ 
\midrule
\texttt{emissions\_2} & What is the carbon intensity (emissions per unit of product or revenue)? \\ 
\midrule
\texttt{emissions\_3} & What are the carbon reduction targets, and what progress has been made? \\ 
\midrule
\texttt{emissions\_4} & How are air pollutants, such as nitrogen oxides and sulfur oxides, managed? \\ 
\midrule
\texttt{resource\_1} & What is the total energy consumption, and what proportion derives from renewable sources? \\ 
\midrule
\texttt{resource\_2} & How is water used, including total consumption and recycling rate? \\ 
\midrule
\texttt{resource\_3} & How efficient is raw material use, and what recycling initiatives are implemented? \\ 
\midrule
\texttt{resource\_4} & How is waste managed, including methods and recycling targets? \\ 
\midrule
\texttt{environment\_1} & What green products or services does the company offer? \\ 
\midrule
\texttt{environment\_2} & How much investment has been made in environmental innovation or R\&D, and what projects are included? \\ 
\midrule
\texttt{environment\_3} & Does the company participate in circular economy initiatives? \\ 
\midrule
\texttt{biodiversity\_1} & What policies are in place to protect biodiversity and natural habitats? \\ 
\midrule
\texttt{biodiversity\_2} & Does the company participate in biodiversity conservation projects, and if so, which ones? \\ 
\midrule
\texttt{biodiversity\_3} & What is the supply chain's impact on land use and the environment? \\ 
\midrule
\texttt{ems\_1} & Does the company have publicly available environmental policies, and if so, what are they? \\ 
\midrule
\texttt{ems\_2} & Has the company obtained environmental certifications such as ISO 14001? If so, which ones? \\ 
\midrule
\texttt{ems\_3} & How transparent is the company regarding environmental data and goals? \\ 
\bottomrule
\end{tabular}
\end{table}

\clearpage
\bibliography{2024_NLP_ref}

@inproceedings{ke2017lightgbm,
  author    = {Ke, Guolin and Meng, Qi and Finley, Thomas and Wang, Taifeng and Chen, Wei and Ma, Weidong and Ye, Qiwei and Liu, Tie-Yan},
  title     = {{LightGBM}: A Highly Efficient Gradient Boosting Decision Tree},
  booktitle = {Advances in Neural Information Processing Systems},
  volume    = {30},
  pages     = {3146--3154},
  year      = {2017},
  publisher = {Curran Associates, Inc.},
  url       = {https://papers.nips.cc/paper/6907-lightgbm-a-highly-efficient-gradient-boosting-decision-tree}
}

@inproceedings{vaswani2017attention,
  author    = {Vaswani, Ashish and Shazeer, Noam and Parmar, Niki and Uszkoreit, Jakob and Jones, Llion and Gomez, Aidan N. and Kaiser, {\L}ukasz and Polosukhin, Illia},
  title     = {Attention is All You Need},
  booktitle = {Advances in Neural Information Processing Systems},
  volume    = {30},
  pages     = {5998--6008},
  year      = {2017},
  publisher = {Curran Associates, Inc.},
  url       = {https://arxiv.org/abs/1706.03762}
}

@inproceedings{devlin2019bert,
  author    = {Devlin, Jacob and Chang, Ming-Wei and Lee, Kenton and Toutanova, Kristina},
  title     = {{BERT}: Pre-training of Deep Bidirectional Transformers for Language Understanding},
  booktitle = {Proceedings of the 2019 Conference of the North {A}merican Chapter of the Association for Computational Linguistics: Human Language Technologies, Volume 1 (Long and Short Papers)},
  pages     = {4171--4186},
  year      = {2019},
  address   = {Minneapolis, Minnesota},
  publisher = {Association for Computational Linguistics},
  doi       = {10.18653/v1/N19-1423},
  url       = {https://aclanthology.org/N19-1423/}
}

@article{johnson2019billion,
  author    = {Johnson, Jeff and Douze, Matthijs and J{\'e}gou, Herv{\'e}},
  title     = {Billion-Scale Similarity Search with {GPUs}},
  journal   = {IEEE Transactions on Big Data},
  volume    = {7},
  number    = {3},
  pages     = {535--547},
  year      = {2021},
  publisher = {IEEE},
  doi       = {10.1109/TBDATA.2019.2921572},
  url       = {https://arxiv.org/abs/1702.08734}
}

@article{liu2019roberta,
  author  = {Liu, Yinhan and Ott, Myle and Goyal, Naman and Du, Jingfei and Joshi, Mandar and Chen, Danqi and Levy, Omer and Lewis, Mike and Zettlemoyer, Luke and Stoyanov, Veselin},
  title   = {{RoBERTa}: A Robustly Optimized {BERT} Pretraining Approach},
  journal = {arXiv preprint arXiv:1907.11692},
  year    = {2019},
  doi     = {10.48550/arXiv.1907.11692},
  url     = {https://arxiv.org/abs/1907.11692}
}

@article{bose2020evolution,
  title={Evolution of ESG reporting frameworks},
  author={Bose, Satyajit},
  journal={Values at work: Sustainable investing and ESG reporting},
  pages={13--33},
  year={2020},
  publisher={Springer},
  url={https://theesgexchange.org/wp-content/uploads/2023/03/Evolution-of-ESG-Reporting-Frameworks.pdf}
}

@inproceedings{guu2020realm,
  author    = {Guu, Kelvin and Lee, Kenton and Tung, Zora and Pasupat, Panupong and Chang, Mingwei},
  title     = {{REALM}: Retrieval-Augmented Language Model Pre-Training},
  booktitle = {Proceedings of the 37th International Conference on Machine Learning},
  series    = {Proceedings of Machine Learning Research},
  volume    = {119},
  pages     = {3929--3938},
  year      = {2020},
  publisher = {PMLR},
  url       = {http://proceedings.mlr.press/v119/guu20a.html}
}

@article{cort2020esg,
  title={ESG standards: Looming challenges and pathways forward},
  author={Cort, Todd and Esty, Daniel},
  journal={Organization \& Environment},
  volume={33},
  number={4},
  pages={491--510},
  year={2020},
  publisher={SAGE Publications Sage CA: Los Angeles, CA},
  doi={10.1177/1086026620945342},
  url={https://www.jstor.org/stable/27001593}
}

@inproceedings{lewis2020retrieval,
  author    = {Lewis, Patrick and Perez, Ethan and Piktus, Aleksandra and Petroni, Fabio and Karpukhin, Vladimir and Goyal, Naman and K{\"u}ttler, Heinrich and Lewis, Mike and Yih, Wen-tau and Rockt{\"a}schel, Tim and Riedel, Sebastian and Kiela, Douwe},
  title     = {Retrieval-Augmented Generation for Knowledge-Intensive {NLP} Tasks},
  booktitle = {Advances in Neural Information Processing Systems},
  volume    = {33},
  pages     = {9459--9474},
  year      = {2020},
  publisher = {Curran Associates, Inc.},
  url       = {https://proceedings.neurips.cc/paper/2020/hash/6b493230205f780e1bc26945df7481e5-Abstract.html}
}

@techreport{gri2021standards,
  author      = {{Global Reporting Initiative}},
  title       = {{GRI} Universal Standards 2021},
  institution = {Global Reporting Initiative},
  year        = {2021},
  address     = {Amsterdam, The Netherlands},
  url         = {https://www.globalreporting.org/standards/}
}

@inproceedings{mehra2022esgbert,
  author    = {Mehra, Srishti and Louka, Robert and Zhang, Yixun},
  title     = {{ESGBERT}: Language Model to Help with Classification Tasks Related to Companies' Environmental, Social, and Governance Practices},
  booktitle = {CS \& IT Conference Proceedings},
  volume    = {12},
  pages     = {183--190},
  year      = {2022},
  doi       = {10.5121/csit.2022.120616},
  url       = {https://arxiv.org/abs/2203.16788}
}

@article{lee2022esg,
  author  = {Lee, Ook and Joo, Hanseon and Choi, Hayoung and Cheon, Minjong},
  title   = {Proposing an Integrated Approach to Analyzing {ESG} Data via Machine Learning and Deep Learning Algorithms},
  journal = {Sustainability},
  volume  = {14},
  number  = {14},
  pages   = {8745},
  year    = {2022},
  publisher = {MDPI},
  doi     = {10.3390/su14148745},
  url     = {https://www.mdpi.com/2071-1050/14/14/8745}
}

@misc{chase2022langchain,
  author       = {Chase, Harrison},
  title        = {{LangChain}},
  year         = {2022},
  howpublished = {\url{https://github.com/langchain-ai/langchain}},
  note         = {Accessed: 2024-12-01},
  url          = {https://github.com/langchain-ai/langchain}
}

@misc{gobel_esg_unstandardized_2022,
  author       = {Julian G{\"o}bel},
  title        = {The impact of unstandardized data on {ESG} reporting},
  howpublished = {Envoria Insights},
  year         = {2022},
  month        = {May},
  url          = {https://envoria.com/insights-news/the-impact-of-unstandardized-data-on-esg-reporting},
  note         = {Accessed: March 27, 2026}
}

@article{rouen2023evolution,
  title={The evolution of ESG reports and the role of voluntary standards},
  author={Rouen, Ethan and Sachdeva, Kunal and Yoon, Aaron},
  journal={Available at SSRN 4227934},
  year={2023},
  url={https://www.hbs.edu/ris/Publication%20Files/23-024_5d9ec300-5c37-4cac-9edb-bcf59650ceb4.pdf}
}

@misc{byrne_esg_history,
  author        = {Dan Byrne},
  title         = {A brief history of {ESG}: {F}rom pioneer to mainstream},
  howpublished = {The Corporate Governance Institute},
  year          = {2023},
  url          = {https://www.thecorporategovernanceinstitute.com/insights/guides/a-brief-history-of-esg-from-pioneer-to-mainstream/},
  note         = {Accessed: May 22, 2024}
}

@article{zou2023esgreveal,
  title={ESGReveal: An LLM-based approach for extracting structured data from ESG reports},
  author={Zou, Yi and Shi, Mengying and Chen, Zhongjie and Deng, Zhu and Lei, ZongXiong and Zeng, Zihan and Yang, Shiming and Tong, HongXiang and Xiao, Lei and Zhou, Wenwen},
  journal={arXiv preprint arXiv:2312.17264},
  year={2023},
  doi={10.48550/arXiv.2312.17264},
  url={https://doi.org/10.48550/arXiv.2312.17264}
}

@thesis{wang2023environmental,
  author      = {Wang, Minzhi Luna},
  title       = {Environmental, Social, and Corporate Governance: A History of {ESG} Standardization from 1970s to the Present},
  institution = {Columbia University},
  department  = {Department of History},
  type        = {Undergraduate Senior Thesis},
  address     = {New York, NY},
  year        = {2023},
  month       = {4},
  url         = {https://sites.asit.columbia.edu/historydept/wp-content/uploads/sites/29/2023/05/Wang-Luna_thesis.pdf},
  note        = {Seminar Advisor: Elizabeth Blackmar; Second Reader: Kimberly Phillips-Fein}
}

@inproceedings{pontes2023leveraging,
  author    = {Pontes, Elvys Linhares and Benjannet, Mohamed and Ming, Lam Kim},
  title     = {Leveraging {BERT} Language Models for Multi-Lingual {ESG} Issue Identification},
  booktitle = {Proceedings of the Multi-Lingual ESG Issue Identification (ML-ESG) Shared Task},
  year      = {2023},
  publisher = {arXiv},
  note      = {arXiv preprint arXiv:2309.02189},
  url       = {https://arxiv.org/abs/2309.02189}
}

@article{ni2023chatreport,
  title={Chatreport: Democratizing sustainability disclosure analysis through llm-based tools},
  author={Ni, Jingwei and Bingler, Julia and Colesanti-Senni, Chiara and Kraus, Mathias and Gostlow, Glen and Schimanski, Tobias and Stammbach, Dominik and Vaghefi, Saeid Ashraf and Wang, Qian and Webersinke, Nicolas and others},
  journal={arXiv preprint arXiv:2307.15770},
  year={2023},
  doi={10.48550/arXiv.2307.15770},
  url={https://doi.org/10.48550/arXiv.2307.15770}
}

@inproceedings{tseng2023dynamicesg,
  title={Dynamicesg: A dataset for dynamically unearthing esg ratings from news articles},
  author={Tseng, Yu-Min and Chen, Chung-Chi and Huang, Hen-Hsen and Chen, Hsin-Hsi},
  booktitle={Proceedings of the 32nd ACM International Conference on Information and Knowledge Management},
  pages={5412--5416},
  year={2023},
  doi={10.1145/3583780.361511},
  url={https://doi.org/10.1145/3583780.361511}
}

@misc{chatpdf2023,
  author       = {{ChatPDF GmbH}},
  title        = {ChatPDF: Specialized AI tool for PDF interaction and summarization},
  year         = {2023},
  howpublished = {Berlin, Germany},
  url          = {https://www.chatpdf.com/},
  note         = {Version 1.4.7 (May 2024). Accessed: 2026-03-27}
}

@misc{chatdoc2023,
  author       = {{ChatDOC}},
  title        = {ChatDOC: AI-powered document analysis and insight extraction},
  year         = {2023},
  month        = mar,
  url          = {https://chatdoc.com/},
  note         = {Initial release March 15, 2023. Terms updated August 2024. Accessed: 2026-03-27}
}

@misc{pdfai2024,
  author       = {{PDF.ai}},
  title        = {PDF.ai: Chat with your PDF documents},
  year         = {2024},
  url          = {https://pdf.ai/},
  note         = {Accessed: 2026-03-27}
}

@misc{wu2024susgengptdatacentricllmfinancial,
      title={SusGen-GPT: A Data-Centric LLM for Financial NLP and Sustainability Report Generation}, 
      author={Qilong Wu and Xiaoneng Xiang and Hejia Huang and Xuan Wang and Yeo Wei Jie and Ranjan Satapathy and Ricardo Shirota Filho and Bharadwaj Veeravalli},
      year={2024},
      eprint={2412.10906},
      archivePrefix={arXiv},
      primaryClass={cs.CL},
      url={https://arxiv.org/abs/2412.10906}, 
}

@inproceedings{xia-etal-2024-using,
    title = "Using Pre-trained Language Model for Accurate {ESG} Prediction",
    author = "Xia, Lei  and
      Yang, Mingming  and
      Liu, Qi",
    booktitle = "Proceedings of the Eighth Financial Technology and
      Natural Language Processing and the 1st Agent AI for Scenario Planning",
    month = "3 " # aug,
    year = "2024",
    address = "Jeju, South Korea",
    url = "https://aclanthology.org/2024.finnlp-2.1",
    pages = "1--22",
}

@misc{khan2024developingretrievalaugmentedgeneration,
      title={Developing Retrieval Augmented Generation (RAG) based LLM Systems from PDFs: An Experience Report}, 
      author={Ayman Asad Khan and Md Toufique Hasan and Kai Kristian Kemell and Jussi Rasku and Pekka Abrahamsson},
      year={2024},
      eprint={2410.15944},
      archivePrefix={arXiv},
      primaryClass={cs.SE},
      url={https://arxiv.org/abs/2410.15944}, 
}

@misc{krantz_esg_history,
  author       = {Tom Krantz},
  title        = {The history of {ESG}: {A} journey towards sustainable investing},
  howpublished = {IBM Think},
  year         = {2024},
  url          = {https://www.ibm.com/think/topics/environmental-social-and-governance-history},
  note         = {Accessed: March 27, 2026}
}

@techreport{lseg2024esg,
  author      = {{London Stock Exchange Group}},
  title       = {Environmental, Social and Governance Scores from {LSEG}},
  institution = {London Stock Exchange Group},
  year        = {2024},
  month       = oct,
  url         = {https://www.lseg.com/content/dam/data-analytics/en_us/documents/methodology/lseg-esg-scores-methodology.pdf}
}

@misc{ding2025euleresgautomatingesgdisclosure,
      title={EulerESG: Automating ESG Disclosure Analysis with LLMs}, 
      author={Yi Ding and Xushuo Tang and Zhengyi Yang and Wenqian Zhang and Simin Wu and Yuxin Huang and Lingjing Lan and Weiyuan Li and Yin Chen and Mingchen Ju and Wenke Yang and Thong Hoang and Mykhailo Klymenko and Xiwei Zu and Wenjie Zhang},
      year={2025},
      eprint={2511.21712},
      archivePrefix={arXiv},
      primaryClass={cs.CL},
      url={https://arxiv.org/abs/2511.21712}, 
}

@misc{yang2025curiousllmelevatingmultidocumentquestion,
      title={CuriousLLM: Elevating Multi-Document Question Answering with LLM-Enhanced Knowledge Graph Reasoning}, 
      author={Zukang Yang and Zixuan Zhu and Xuan Zhu},
      year={2025},
      eprint={2404.09077},
      archivePrefix={arXiv},
      primaryClass={cs.CL},
      url={https://arxiv.org/abs/2404.09077}, 
}

@inproceedings{mousavian-anaraki-etal-2025-automatic,
    title = "Automatic {GRI}-{SDG} Annotation and {LLM}-Based Filtering for Sustainability Reports",
    author = "Mousavian Anaraki, Seyed Alireza and
      Croce, Danilo and
      Basili, Roberto",
    editor = "Bosco, Cristina and
      Jezek, Elisabetta and
      Polignano, Marco and
      Sanguinetti, Manuela",
    booktitle = "Proceedings of the Eleventh Italian Conference on Computational Linguistics (CLiC-it 2025)",
    month = sep,
    year = "2025",
    address = "Cagliari, Italy",
    publisher = "CEUR Workshop Proceedings",
    url = "https://aclanthology.org/2025.clicit-1.73",
    pages = "775--784",
}

@misc{zhang2025mmesgbenchpioneeringmultimodalunderstanding,
      title={MMESGBench: Pioneering Multimodal Understanding and Complex Reasoning Benchmark for ESG Tasks}, 
      author={Lei Zhang and Xin Zhou and Chaoyue He and Di Wang and Yi Wu and Hong Xu and Wei Liu and Chunyan Miao},
      year={2025},
      eprint={2507.18932},
      archivePrefix={arXiv},
      primaryClass={cs.MM},
      url={https://arxiv.org/abs/2507.18932}, 
}

@inproceedings{He_2025,
   title={ESGenius: Benchmarking LLMs on Environmental, Social, and Governance (ESG) and Sustainability Knowledge},
   url={http://dx.doi.org/10.18653/v1/2025.emnlp-main.739},
   DOI={10.18653/v1/2025.emnlp-main.739},
   booktitle={Proceedings of the 2025 Conference on Empirical Methods in Natural Language Processing},
   publisher={Association for Computational Linguistics},
   author={He, Chaoyue and Zhou, Xin and Wu, Yi and Yu, Xinjia and Zhang, Yan and Zhang, Lei and Wang, Di and Lyu, Shengfei and Xu, Hong and Xiaoqiao, Wang and Liu, Wei and Miao, Chunyan},
   year={2025},
   pages={14623–14664} }

@article{anaraki2025large,
  title={Large language models for sustainability reporting: A systematic review and research agenda},
  author={Anaraki, Seyed Alireza Mousavian and Croce, Danilo and Basili, Roberto},
  journal={Sustainable Futures},
  volume={10},
  pages={101494},
  year={2025},
  publisher={Elsevier},
  doi={10.1016/j.sftr.2025.101494},
  url={https://doi.org/10.1016/j.sftr.2025.101494}
}

@misc{ecoactive_esg_challenges_2025,
  author       = {{EcoActive}},
  title        = {The 5 {M}ain {C}hallenges of {ESG} {R}eporting and {B}est {P}ractices},
  howpublished = {EcoActive Blog},
  year         = {2025},
  month        = {February},
  url          = {https://ecoactivetech.com/the-5-main-challenges-of-esg-reporting-and-best-practices/},
  note         = {Accessed: March 27, 2026}
}

@misc{hoang2026esgreportinglifecyclemanagement,
      title={ESG Reporting Lifecycle Management with Large Language Models and AI Agents}, 
      author={Thong Hoang and Mykhailo Klymenko and Xiwei Xu and Shidong Pan and Yi Ding and Xushuo Tang and Zhengyi Yang and Jieke Shi and David Lo},
      year={2026},
      eprint={2603.10646},
      archivePrefix={arXiv},
      primaryClass={cs.SE},
      url={https://arxiv.org/abs/2603.10646}, 
}

\end{document}